\documentclass[review]{elsarticle}

\usepackage[bookmarks=false]{hyperref}
\usepackage{epstopdf}
\usepackage{subfig}
\usepackage[percent]{overpic}
\usepackage{amsfonts}
\usepackage{amsmath}
\usepackage{amssymb}
\usepackage{mathrsfs}
\usepackage{fixltx2e}
\newcommand{\argmin}[1]{\underset{#1}{\operatorname{arg}\,\operatorname{min}}\;}

\journal{Journal of Structural Biology}

\begin{document}

\begin{frontmatter}

\title{Denoising and Covariance Estimation of Single Particle Cryo-EM Images}
%
\author{Tejal Bhamre}
\address{Department of Physics, Princeton University, Jadwin Hall, Washington Road, Princeton, NJ 08544-0708, USA}

\author{Teng Zhang}
\address{Department of Mathematics, University of Central Florida, 4393 Andromeda Loop N, Orlando, FL 32816-8007, USA}

\author{Amit Singer}
\address{Department of Mathematics and PACM, Princeton University, Fine Hall, Washington Road, Princeton, NJ 08544-1000, USA}


%
%
%
%
\begin{abstract}
The problem of image restoration in cryo-EM entails correcting for the effects of the Contrast Transfer
Function (CTF) and noise. Popular methods for image restoration include
`phase flipping', which corrects only for the Fourier phases but not amplitudes, and Wiener filtering, which
requires the spectral signal to noise ratio. We propose a new image restoration method
which we call `Covariance Wiener Filtering' (CWF).
In CWF, the covariance matrix of the projection images is used within the 
classical Wiener filtering framework for solving the image restoration 
deconvolution problem. Our estimation procedure for the covariance matrix is new 
and successfully corrects for the CTF.   
We demonstrate the efficacy of CWF by applying it to restore both simulated and experimental cryo-EM images.
Results with experimental datasets demonstrate that CWF provides a good way to 
evaluate the particle images and to see what the dataset contains even without 2D 
classification and averaging.

\end{abstract}


\end{frontmatter}

\section{Introduction}
Single particle reconstruction (SPR) using cryo-electron microscopy (cryo-EM) is 
a rapidly advancing technique for determining the structure of biological
macromolecules at near-atomic resolution directly in their native state, without any need for crystallization \cite{cryoem_rev, rev2, nogales, sigworth_rev, kuhlbrandt}. In SPR, 3D reconstructions are estimated by combining multiple noisy 2D tomographic projections
of macromolecules in different unknown orientations.

The acquired data consists of multiple micrographs from which particle images 
are extracted in the first step of the computational pipeline. Next, the images are grouped together
by similarity in the 2D classification and averaging step \cite{jane_classav,park}. Class averages can be used
to inspect the underlying particles, and
to estimate viewing angles and form a low resolution ab-initio 3D model. Subsequently, this 3D model 
is refined to high resolution, and 3D classification might be performed as well. 

In this paper we propose an image restoration method that provides a way for 
visualizing the particle images without performing any 2D classification. 
While noise reduction is achieved in 2D classification by averaging together 
different particle images, our method operates on each image separately, and 
performs contrast transfer function (CTF) correction and denoising in a single 
step.

Existing image restoration techniques (for denoising and CTF correction) can be 
broadly categorized into two kinds of approaches \cite{Penczek_image}.
The first is an approach known as `phase flipping', which involves 
flipping the sign of the Fourier coefficients at frequencies for which the CTF 
is negative.
Consequently, phase flipping restores the correct phases of the Fourier 
coefficients, but ignores the effect of the CTF on the
amplitudes. Phase flipping preserves the noise statistics and is easy to 
implement, leading to its widespread usage in several cryo-EM software packages.
However, it is suboptimal because it does not restore the correct
Fourier amplitudes of the images. The second commonly used approach is Wiener 
filter based restoration, to which we refer here as traditional Wiener filtering (TWF). 
Wiener filtering takes into account both the phases and amplitudes
of the Fourier coefficients, unlike phase flipping. However, calculation of the 
Wiener filter coefficients requires prior estimation
of the spectral signal to noise ratio (SSNR) of the signal, which by itself is a challenging 
problem. It is therefore customary to either
treat the SSNR as a precomputed constant as in the software package SPIDER 
\cite{spider}, or to
apply Wiener filtering only at later stages of the 3D reconstruction pipeline 
when the noise level is sufficiently low, such as in
EMAN2 \cite{eman2}. It is also possible to use a combination of the two 
approaches, by first phase flipping the 2D images,
and later correct only for the amplitudes in the 3D reconstruction step, as in 
IMAGIC \cite{imagic, imagic2}. Despite its simplicity,
there are several drawbacks to TWF. First, it cannot restore information at the 
zero 
crossings of the CTF. Second, it requires estimation of the SSNR. Third, it is
restrictive to the Fourier basis which is a fixed basis not adaptive to the 
image dataset.

We refer to our proposed method as Covariance
Wiener Filtering (CWF). CWF consists
of first estimating the CTF-corrected covariance matrix of the underlying clean 
2D projection images, followed by application of the Wiener filter
to denoise the images. Unlike phase flipping, CWF takes into 
account both the phases and magnitudes
of the images. Moreover, unlike TWF that always operates in the data-independent 
Fourier domain, CWF is performed in the data-dependent basis of principal 
components (i.e., eigenimages).
Crucially, CWF can be applied at preliminary
stages of data processing on raw 2D particle images. The resulting denoised 
images can be used for an early inspection of the dataset,
to identify the associated symmetry, and to eliminate `bad' particle images 
prior to 2D classification and 3D reconstruction. Additionally, the estimation 
of the
2D covariance matrix is itself of interest, for example, in Kam's approach for 
3D reconstruction \cite{kam1980, or}.

The paper is organized as follows: sections \ref{sec:model} and 
\ref{sec:covest_col} detail the estimation of the covariance matrix for two different noise models, first for the
simpler model of white noise, and second for the more realistic model of colored noise.
In section \ref{sec:fb_basis} we discuss the steerability
property of the covariance matrix \cite{ffbspca}. 
The associated deconvolution problem
is solved to obtain denoised images using the estimated covariance matrix in section \ref{sec:wf}.
Finally in section \ref{sec:results},
we demonstrate CWF in a number of numerical experiments, with both simulated and experimental datasets.
We obtain encouraging results for experimental datasets, in particular, those acquired with the modern
direct electron detectors. Image features are clearly observed after CWF denoising. For reproducibility,
the MATLAB code for CWF and its dependencies are available in the open source
cryo-EM toolbox ASPIRE at \url{www.spr.math.princeton.edu}. The script \textit{cwf\_script.m},
calls the main function \textit{cwf.m}.

\section{Methods}
The first 
step of CWF is estimation of the
covariance matrix of the underlying clean images, to which we refer as
the population covariance. The second step of CWF is 
solving a deconvolution problem to recover
the underlying clean images using the estimated covariance. In the rest 
of this section,
we describe these steps in detail.

\subsection{The Model}
\label{sec:model}
The image formation model in cryo-EM under the linear, weak phase 
approximation \cite{Frankbook} is given by 

\begin{equation}
 y_i = a_i \ast x_i + \epsilon_i, \quad i=1,2,\ldots,n
\label{eqn:model}
\end{equation}
where $n$ is the number of images, $\ast$ denotes the convolution operation, $y_i$ is the noisy, CTF filtered $i$'th image in real
space, $x_i$ is the underlying clean projection image in 
real space, 
$a_{i}$ is the point spread function of the microscope that
convolves with the clean image in real space, and $\epsilon_i$ is 
additive Gaussian noise that corrupts the image, for each $i$. Taking the Fourier transform of 
eqn. \ref{eqn:model} gives

\begin{equation}
 Y_i = A_i X_i + \xi_i, \quad i=1,2,\ldots,n
\label{eqn:model_f}
\end{equation}
where $Y_i$, $X_i$ and $\xi_i$ are now in Fourier space. $A_i$ is 
a diagonal operator, whose diagonal consists of
the Fourier transform of the point spread function,
and is also commonly known as the CTF. The CTF modulates the
phases and the amplitudes of the Fourier coefficients of the image, and contains
numerous zero crossings that correspond
to frequencies at which no information is obtained.
Any image restoration technique that aims to completely correct for the CTF
must therefore correctly restore both the phases and the amplitudes.
The zero crossings make CTF correction
challenging since it cannot be trivially inverted. 
In experiments, different
groups of images are acquired at different defocus values, in the hope that 
information that is lost from one group could be
recovered from another group that has different zero crossings. In the experimental
datasets used in this paper, the number of images per defocus group typically ranges from 
$50$ to $1000$.

In our statistical model, the Fourier transformed clean images $X_1,\dots,X_n$ 
(viewed, for mathematical convenience, as vectors in $\mathbb{C}^p$, where $p$ is the number of pixels) 
are assumed to be independent,
identically distributed (i.i.d.) samples from a distribution with mean $\mathbb{E}[\textbf{X}]=\mu$
and covariance $\mathbb{E}[(\textbf{X}-\mu)(\textbf{X}-\mu)^T]=\Sigma$. 
Since the clean images are two-dimensional projections of the three-dimensional molecule in different orientations, 
the distribution of $\textbf{X}$ in our model is determined by the three-dimensional structure, the distribution of orientations, 
the varying contrast due to changes in ice thickness, and structural variability, all of course unknown at this stage. 
The covariance matrix $\Sigma$ therefore represents the overall image variability due to these determinants.
While these model assumptions do not necessarily hold in reality \cite{sorzano1, sorzano2}, they simplify the analysis and, as will be shown later lead to excellent denoising.
Quoting George Box, ``All models are wrong but some are useful" \cite{box}.

Our denoising scheme requires $\mu$ and
$\Sigma$. Since these quantities are not readily given, we estimate them from the noisy images themselves as follows.
For simplicity, we first assume that the noise in our model is additive 
white
Gaussian noise such that $\xi_i \sim \mathcal{N} (0,\sigma^2 I_{p \times p})$ 
in eqn. \ref{eqn:model_f} are i.i.d. The white noise assumption is later replaced by that 
of the more realistic colored noise. First, notice from eqn. \ref{eqn:model_f} it follows that

\begin{equation}
\mathbb{E}[\textbf{Y}_i]=A_i \mathbb{E}[\textbf{X}_i], \quad i=1,2,\ldots,n.
\label{eqn:exp_y}
\end{equation}
So,
\begin{equation}
\begin{aligned}
\mathbb{E}[(\textbf{Y}_i-\mathbb{E}[\textbf{Y}_i])(\textbf{Y}_i-\mathbb{E}[\textbf{Y}_i])^T] 
&= \mathbb{E} [A_i(\textbf{X}_i-\mu)(\textbf{X}_i-\mu)^T A_i^T] + \sigma^2I \\
&=  A_i \Sigma A_i^T + \sigma^2I .
\end{aligned}
\label{eqn:expectation_eq}
\end{equation}
Eqn. \ref{eqn:expectation_eq} relates the second order statistics of the noisy images with the 
population covariance $\Sigma$ of the clean images, based on which we can 
estimate $\Sigma$.

Next, we construct estimators for the mean $\mu$ and population covariance 
$\Sigma$ using eqn. \ref{eqn:exp_y} and \ref{eqn:expectation_eq}.
The mean $\mu$ of the dataset can be estimated 
as the solution to
a least squares problem

\begin{equation}
 \hat\mu = \argmin{\mu} \sum_{i=1}^n||(Y_i-A_i\mu)||_2^2 + 
\lambda||\mu||_2^2
\label{eq:ls_mean}
\end{equation}
where $\lambda \geq 0$ is a regularization parameter. The solution to 
\ref{eq:ls_mean} is explicitly
\begin{equation}
 \hat\mu = (\sum_{i=1}^n A_i^T A_i + \lambda I)^{-1}(\sum_{i=1}^n 
A_i^T Y_i).
\label{eq:ls_mean_sol}
\end{equation}
The population covariance $\Sigma$ can be estimated as 
\begin{equation}
\begin{aligned}
\hat\Sigma 
&= \argmin{\Sigma} \sum_{i=1}^n || (Y_i - \mathbb{E}[\textbf{Y}_i]) (Y_i - \mathbb{E}[\textbf{Y}_i])^T
- (A_i \Sigma A_i^T + \sigma^2 I)||_F^2 \\
&= \argmin{\Sigma} \sum_{i=1}^n || A_i\Sigma A_i^T + \sigma^2 I - C_i  ||_F^2 
\end{aligned}
\label{eqn:ls1}
\end{equation}
where $C_i=(Y_i - A_i \mu) (Y_i - A_i \mu)^T$ and $||.||_F$ is the Frobenius matrix norm. 
The estimators $\hat \mu$ and $\hat \Sigma$ can be shown to be consistent in the large sample limit
$n \to \infty$, similar to the result in Appendix B of \cite{gene}. 

To ensure that the estimated
covariance is positive semidefinite (PSD), we project it onto the space of 
PSD matrices by computing
its spectral decomposition and retaining only the non negative eigenvalues (and their corresponding eigenvectors).
To solve eqn.\ \ref{eqn:ls1},
we differentiate the objective function with respect to $\Sigma$ 
and set the derivative to zero. This yields

\begin{equation}
\begin{aligned}
\sum_{i=1}^n  A_i^T  A_i \hat \Sigma A_i^T A_i
&= \sum_{i=1}^n A_i^T C_i A_i - \sum_{i=1}^n \sigma^2 A_i^T A_i 
\end{aligned}
\label{eqn:normal_white}
\end{equation}
Eqn.\ \ref{eqn:normal_white} defines a system of linear equations for the elements of the matrix $\hat \Sigma$.
However, direct inversion of this linear system is slow and computationally impractical for large image sizes.
Notice that eqn.\ \ref{eqn:normal_white} can be written as 
\begin{equation}
\begin{aligned}
L(\hat\Sigma) 
&=  B 
\label{eqn:cg}
\end{aligned}
\end{equation}
where $L:\mathbb{R}^{p\times p} \to \mathbb{R}^{p\times p}$ is the linear operator acting on $\hat{\Sigma}$ defined by the left hand side of eqn. \ref{eqn:normal_white}, and $B$ is the right hand side.
Since applying $L$ only involves matrix multiplications, it can be computed fast, and
the conjugate gradient method is employed to efficiently compute $\hat \Sigma$ instead of direct inversion,
similar to how it is used in \cite{joakim}.

Notice that $L(\hat{\Sigma})$ is a PSD matrix whenever $\hat{\Sigma}$ is PSD (as 
a sum of PSD matrices), while $B$ may not necessarily be PSD due to finite 
sample fluctuations (i.e., $n$ is finite). It is therefore natural to project 
$B$ onto the cone of PSD matrices. This amounts to computing the spectral 
decomposition of $B$ and setting all negative eigenvalues to 0, which is an 
instance of eigenvalue thresholding.

We now describe an alternate eigenvalue thresholding procedure, better suited to 
cases in which the number of images $n$ is not exceedingly large. To that end, 
we first analyze the matrix $B$ when $X_i=0$ for all $i$, i.e., the input images 
are white noise images containing no signal. Let 
\begin{equation}
M = \sum_{i=1}^n A_i^T C_i A_i = \sum_{i=1}^n A_i^T Y_i Y_i^T A_i.
\end{equation}
Then, $\mathbb{E}[M] = \sigma^2 \sum_{i=1}^n A_i^T A_i$
and $B = M - \mathbb{E}[M]$. Let $S = (\mathbb{E}[M])^{1/2}$, i.e. 
$S$ is PSD and $\mathbb{E}[M]=S^2$.
Then multiplying both sides of eqn. \ref{eqn:cg} with $S^{-1}$ we get

\begin{equation}
 S^{-1} L(\hat\Sigma)  S^{-1} = S^{-1}(M - \mathbb{E}[M]) S^{-1} = S^{-1} M S^{-1} - I .
\label{eqn:pop1}
\end{equation}
$S^{-1}MS^{-1}$ can be viewed as a sample covariance matrix of
$n$ vectors in $ \mathbb{R}^p$ whose population covariance is the identity matrix.
When $p$ is fixed and $n$ goes to infinity, all eigenvalues of $S^{-1}MS^{-1}$ converge to $1$.
In practice, however, $n$ and $p$ are often comparable. In the limit $p, n \to 
\infty$ and $p/n \to \gamma$ with $0 < \gamma < \infty$, the limiting spectral 
density of the eigenvalues converges to the Mar\v{c}enko Pastur (MP) distribution \cite{marcenko},
given by

\begin{equation}
MP(x) = \frac{1}{2\pi}\frac{\sqrt{(\gamma_+ - x)(x - \gamma_-)}}{\gamma x}1_{[\gamma_-,\gamma_+]}, \quad \gamma_{\pm} =(1 \pm \sqrt{\gamma})^2 
\end{equation}
for $\gamma \leq 1$. It is therefore expected that $S^{-1}MS^{-1}$ would have eigenvalues 
(considerably) larger than $1$, even in the pure white noise case. These large 
eigenvalues should not be mistakingly attributed to signal. In the case of 
images containing signal (plus noise), eigenvalues corresponding to the signal 
can only be detected if they reside outside of the support of the MP 
distribution. We use the method of \cite{knadler} to determine the number of eigenvalues 
corresponding to the signal. We then apply the operator norm eigenvalue 
shrinkage procedure (see \cite{donoho}) to those eigenvalues, while setting all other eigenvalues to 
$0$. We then use the conjugate gradient method \footnote{While $L$ in eqn.\ 9 is PSD, the new effective operator in the LHS of eqn.\ 11 is not necessarily PSD in general.
In order to use conjugate gradient, we solve the system $S^{-1}L(S^{-1}\Sigma_S S^{-1})S^{-1} = S^{-1}MS^{-1}-I$, where
$\Sigma_S = S \Sigma S$, in which the operator acting on $\Sigma_S$ in the LHS is PSD. $\Sigma$ is then obtained
from the estimated $\Sigma_S$.
} to solve eqn. \ref{eqn:pop1} for $\hat{\Sigma}$, with the right hand side 
replaced with its shrinkage version. We observed in numerical simulations (see Fig. \ref{fig:shrinkage}) that this procedure
typically outperforms other shrinkage methods in terms of the accuracy of the estimated covariance matrix.

\subsection{Covariance Estimation with Colored noise}
\label{sec:covest_col}
So far, we assumed additive white Gaussian noise in the image formation 
process. In reality, the noise in experimental images
is colored. That is, in the image formation model in eqn.\ \ref{eqn:model_f},
$\xi_i$ is additive colored Gaussian noise. 
We preprocess the images in order to ``whiten'' the noise.
The noise power spectrum can be estimated, for example, using the pixels in the corners of the noisy projection images.
To do this, we first estimate using correlograms the 2D autocorrelation
of the corner pixels of the images which contain mostly noise and no signal. These corner
pixels are used to estimate the 1D autocorrelation, which is then extended to populate the 2D isotropic
autocorrelation. We then calculate the Fourier transform of the 2D autocorrelation, 
which is the 2D power spectrum of noise.
The noisy projection images in 
Fourier space are multiplied element-wise by the inverse of the estimated 
power spectral density, also called the whitening filter, so that the noise in the resulting images 
is approximately white.
Let $W$ be the ``whitening'' filter, such that

\begin{equation}
 WY_i=WA_{i}X_i + W\xi_i, \quad i=1,2,\ldots,n
 \label{eqn:col_model}
\end{equation}
and $W \xi_i \sim \mathcal{N}(0,\sigma^2 I)$. 

Eqn.\ \ref{eqn:col_model} is reminiscent of eqn.\ \ref{eqn:model_f}. It is tempting to define a new 
``effective'' CTF as $WA_{i}$ and estimate
$\Sigma$ following the same procedure as in the case of white noise. However, 
the linear system akin to eqn.\ \ref{eqn:normal_white} for this case
is ill-conditioned due to the product of $W$ with the CTF, and it 
takes a large number of iterations for conjugate gradient 
to converge to the desired solution. Instead, we seek an approach in which the 
linear system to solve is well conditioned as that
in the case of white noise.
Since the CTF's $A_i,\ i=1,2, \ldots n$ 
and the whitening filter $W$ are diagonal operators in the Fourier basis, they 
commute, and eqn.\ \ref{eqn:col_model} becomes
\begin{equation}
WY_i=A_{i}WX_i + W\xi_i, \quad i=1,2,\ldots,n.
\label{eqn:model_col}
\end{equation}
We therefore 
absorb $W$ into $X_i$, and estimate the matrix $\Sigma_W = 
W\Sigma W^T$ (the population covariance of $W \textbf{X}$) using the same procedure as before.
The population covariance $\Sigma$ is then estimated as

\begin{equation}
\hat{\Sigma} = W^{-1} \hat{\Sigma}_W (W^T)^{-1} .
\end{equation}

\subsection{Fourier-Bessel Steerable PCA}
\label{sec:fb_basis}
The population covariance matrix $\Sigma$ must be invariant under in-plane 
rotation of the projection images, therefore it is block diagonal in any steerable basis in which the 
basis elements are outer products of radial functions and angular Fourier modes. 
Following \cite{ffbspca}, we 
choose to represent the images in a Fourier-Bessel basis and it suffices to 
estimate each diagonal block $\Sigma^{(k)}$, corresponding to the angular frequency $k$, separately. 
The Fourier-Bessel basis \cite{ffbspca} consists of $p_k$ basis functions (that satisfy
the sampling criterion) for each angular frequency $k$, where $p_k$ decreases with increasing $k$.
The matrix $\Sigma^{(k)}$ is thus of size $p_k \times p_k$.

An important property
of the CTF's $A_i$ and the whitening filter $W$ is that they are radially isotropic \footnote{In the case of
astigmatism, where the CTF deviates slightly from radial isotropy, this is a 
good approximation to obtain low resolution
denoised images.}. 
Therefore, the CTF's and the whitening filter
are also block diagonal in the Fourier Bessel basis. Eqn. 
\ref{eqn:normal_white} (and its analog in the case of colored noise)
is hence solved separately for each $k$ to estimate $\Sigma^{(k)}$.

\subsection{Wiener Filtering}
\label{sec:wf}
The estimated covariance is
further used to solve the associated deconvolution problem
in eqn. \ref{eqn:model_f} using Wiener filtering. The result is a denoised, CTF 
corrected image for each noisy, CTF affected
measurement $Y_i$ for $i=1,2,\ldots n$. We
estimate $X_i$ in the white noise model using the Wiener 
filtering procedure as
\begin{equation}
\hat X_i = (I-H_iA_{i})\hat\mu + H_iY_i 
\end{equation}
where $H_i = \hat \Sigma A_{i}^T ( A_{i} \hat \Sigma A_{i}^T + \sigma^2 
I)^{-1} $ is the linear Wiener filter \cite{wiener_deriv}. In the case of colored noise,
\begin{equation}
\hat X_i = (I-H_iWA_{i})\hat\mu + H_iY_i 
\end{equation}
with $H_i = \hat \Sigma A_{i}^T W^T (W A_{i} \hat \Sigma A_{i}^T W^T 
+ \sigma^2 I)^{-1}$. Since the estimated covariance
is block-diagonal in the Fourier Bessel basis, the Wiener filtering
procedure is applied to the Fourier Bessel coefficients of the noisy 
images $Y_i$ for each angular frequency $k$ separately. The denoised 
Fourier Bessel expansion coefficients
are used to reconstruct denoised images in Fourier space that are inverse Fourier 
transformed to acquire images in real space on a Cartesian grid.

\subsection{Computational Complexity}
\label{sec:complexity}
In practice, instead of each image being affected by a distinct CTF, all images 
within a given defocus group have the same CTF. So, given $D$ defocus groups 
with $d_i$
images in group $i$, one can equivalently minimize the objective function
 $\sum_{i=1}^D d_i ||(A_i \Sigma A_i^T + \sigma^2 I) -
 \sum_{j=1}^{d_i} \frac{1}{d_i}(Y_{i_j} - \mathbb{E}[\textbf{Y}_{i_j}]) (Y_{i_j} - 
\mathbb{E}[\textbf{Y}_{i_j}])^T
||_F^2 $ in eqn.\ \ref{eqn:ls1} (here $A_i$ denotes the CTF of the $i$'th defocus group, and $i_j$ index images in that group). 
As a result, the sums in eqn.\ \ref{eqn:normal_white} range 
from $1$ to $D$ instead of from $1$ to $n$, thereby reducing the computational cost 
of some operations.
For images of size $L \times L$, estimating the mean using eqn. 
\ref{eq:ls_mean_sol} takes $O(nL^2)$ (since $A_i$ is diagonal
in the Fourier basis for each $i$). Computing the Fourier Bessel expansion 
coefficients takes $O(nL^3)$, as detailed
in \cite{ffbspca}. When solving the linear system in eqn. \ref{eqn:normal_white}
to estimate each $\Sigma^{(k)}$ separately, the matrices in eqn. 
\ref{eqn:normal_white} are of size $p_k \times p_k$.
It is shown in \cite{ffbspca} that $\sum_k p_k=O(L^2)$, $\sum_k p_k^2=O(L^3)$, 
and $\sum_k p_k^3=O(L^4)$.
While solving eqn. \ref{eqn:cg} using conjugate gradient for a given angular 
frequency, computing the 
action of the linear operation $L$ on $\Sigma^{(k)}$ takes
$O(Dp_k^3)$ per iteration,
while computing $B$ takes $O(Dp_k^3+np_k^2)$. Thus, each iteration of 
conjugate gradient takes
$O(D \sum_k p_k^3)$, that is,
$O(DL^4)$ and there is also a one time computation of $O(nL^3)$.
Wiener filtering the Fourier Bessel coefficients of an image for a given 
angular frequency $k$ takes $O(p_k^2)$. So the overall
complexity for Wiener filtering the coefficients of all images is $O(nL^3)$. 
In summary, the overall complexity for CWF is 
$O(TDL^4 + nL^3)$, where $T$ is the number of conjugate gradient iterations.

\section{Results}
\label{sec:results}
In this section, we apply our algorithm to synthetic and experimental datasets 
to obtain denoised images. All algorithms are
implemented in the UNIX environment, on a machine with 60 cores,
running at 2.3 GHz, with total RAM of 1.5TB. We perform numerical experiments 
with
(i) a synthetic dataset with additive white and colored Gaussian noise and (ii) 
four experimental
datasets, two of which were acquired with older detectors, and the other two 
with state-of-the-art
direct electron detectors. For all the experimental datasets, the corresponding
estimated CTF parameters were provided with the dataset. For all simulations,
we use centered projection images. The algorithm does not require centered images. 
However, having non-centered images would result in an additional
'blurring' effect in the denoised images.

\subsection{Simulated Noisy Dataset with White Noise}
\label{sec:whitenoise}
For the first experiment with simulated data, we construct a synthetic dataset 
by modeling the image formation process in 
cryo-EM. The synthetic dataset is prepared from the 3D structure of the P. 
falciparum 80S ribosome bound to E-tRNA, available on the Electron
Microscopy Data Bank (EMDB) as EMDB-6454. We first generate clean 2D projection 
images starting from a 3D volume, at directions sampled uniformly
over the sphere, and then corrupt the generated clean projection images with 
different CTF's and additive white Gaussian 
noise. The projection images are divided into $10$ defocus groups, with the 
defocus value
ranging from $1\mu m$ to $4\mu m$. The B-factor of the decay envelope was 
chosen 
as $10\AA^2$, the amplitude contrast as $7\%$,
the voltage as $300$kV, and the spherical aberration as $2$mm. To ensure that 
the denoising quality of CWF is robust to the mean estimation of the dataset, 
the regularization parameter $\lambda$ in the least squares mean estimation in eqn. \ref{eq:ls_mean_sol} was 
fixed at 1 for all the experiments
described here.

\begin{figure}[]
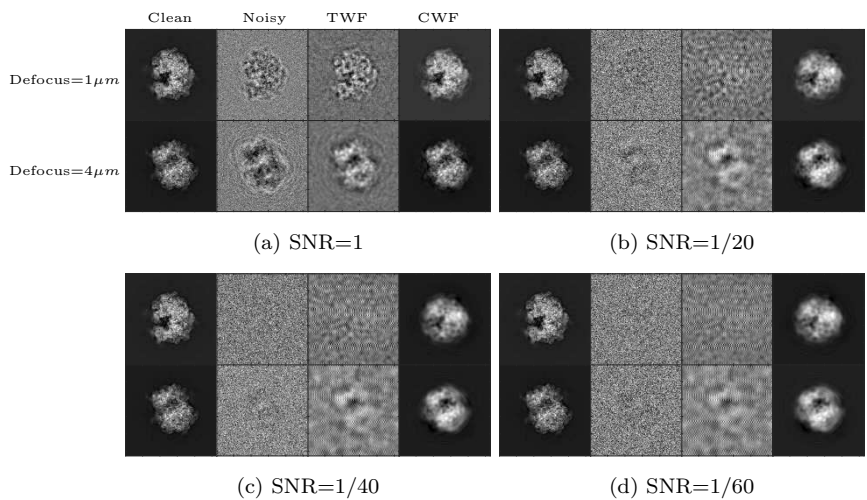

\centering
\subfloat[SNR=$1$]{\begin{overpic}[width=0.4\linewidth]{compare_ims_snr1by1_6454.eps}
\put(6,52){\tiny Clean}
\put(32,52){\tiny Noisy}
\put(55,52){\tiny TWF}
\put(80,52){\tiny CWF}
\put(-32,35){\tiny Defocus=$1\mu m$}
\put(-32,10){\tiny Defocus=$4\mu m$}
\end{overpic}%
}
\vspace{-1mm}
\subfloat[SNR=$1/20$]{\begin{overpic}[width=0.4\linewidth]{compare_ims_snr1by20_6454.eps}%
\end{overpic}%
\label{}}
\vspace{-1mm}
\subfloat[SNR=$1/40$]{\begin{overpic}[width=0.4\linewidth]{compare_ims_snr1by40_6454.eps}%
\end{overpic}%
\label{}}
\vspace{-3mm}
\subfloat[SNR=$1/60$]{\begin{overpic}[width=0.4\linewidth]{compare_ims_snr1by60_6454.eps}%
\end{overpic}%
\label{}}
\caption{\textbf{Synthetic white noise}: A comparison of the denoising results of 
traditional Wiener filtering (TWF) and CWF for the 
synthetic dataset prepared from EMDB-6454, the P. falciparum 80S ribosome bound 
to E-tRNA. The dataset consists of 10000 images
of size 105$\times$105, which are divided into 10 defocus groups, with the 
defocus value ranging from 1$\mu m$ to 4$\mu m$. The two rows in each subfigure 
correspond to two clean
images belonging to different defocus groups;
the first one
belongs to the group with the smallest defocus value of 1$\mu m$, 
while the second image belongs to the group with the
largest defocus value of 4$\mu m$.  
}
\label{fig:ims_6454}
\end{figure}

Figure \ref{fig:ims_6454} shows the results of denoising 
raw, CTF-affected noisy images with CWF and
TWF at various levels of the SNR. We have used the EMAN2 \cite{eman2}
implementation of TWF (note that we perform phase flipping followed 
by TWF only on the raw images in EMAN2, and not on averages). The SNR used here is defined relative to the 
CTF affected images that constitute the clean signal, and is calculated
as an average value for the entire dataset. Using 20 cores, calculating the Fourier Bessel
coefficients took 79 seconds while covariance estimation and Wiener filtering together
took 6 seconds in the experiment with SNR$=1/60$. 

It is seen that TWF works very well at high SNR ($\geq1$), but deteriorates 
at lower SNR's as expected. Note that the denoising results of TWF depend 
strongly on the defocus value. 
The location of the zeros in the CTF is such that images 
corresponding to high defocus values preserve low frequency information, while 
images corresponding to low defocus values
retain more high frequency information. With CWF, there is no such strong 
dependence on the defocus value, since the covariance matrix is estimated
using information from all defocus groups.

\begin{figure}[]
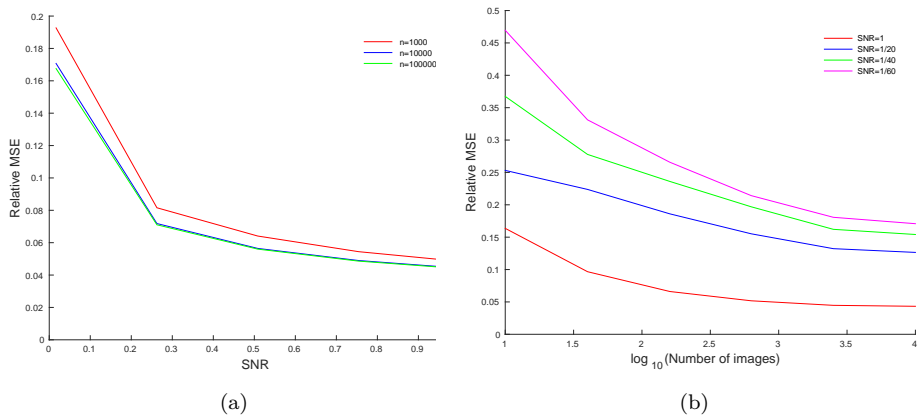

\centering
\subfloat[]{\begin{overpic}[width=0.5\linewidth]{mse_snr_6454.eps}%
\end{overpic}%
\label{fig:mse_snr}}
\subfloat[]{\begin{overpic}[width=0.5\linewidth]{mse_nims_6454.eps}
\end{overpic}
\label{fig:mse_nims}}
\caption{(a) \textbf{Relative MSE versus the SNR, for a fixed number of images}: 
The 
relative MSE of the denoised images as a function of the SNR, for synthetic 
data 
generated using EMDB-6454. The 
MSE reported here is averaged over all images. \textit{n} denotes the number of images 
used in the experiment.(b) \textbf{Relative MSE versus the number of images, for a fixed SNR}: 
The 
relative MSE of the denoised images as a function of the number of images, for 
synthetic data generated using EMDB-6454. The 
MSE reported here is averaged over all images. }
\end{figure}

Figure \ref{fig:mse_snr} shows the relative MSE of denoised images as a function
of the SNR of the dataset. The MSE (norm of the difference between the denoised image and the original, clean image) shown here
corresponds to the same range of SNR's (from $1/60$ to $1$) as in Figure 
\ref{fig:ims_6454}. Figure \ref{fig:mse_nims} shows the relative MSE of the 
denoised images as a 
function of the number of images used to estimate the covariance in the experiment.
The covariance estimation improves as the number of images in the 
dataset increases, and so the denoising is also expected
to improve, as seen from Figure \ref{fig:mse_nims}.

\begin{figure}
\centering

\includegraphics[width=0.8\linewidth]{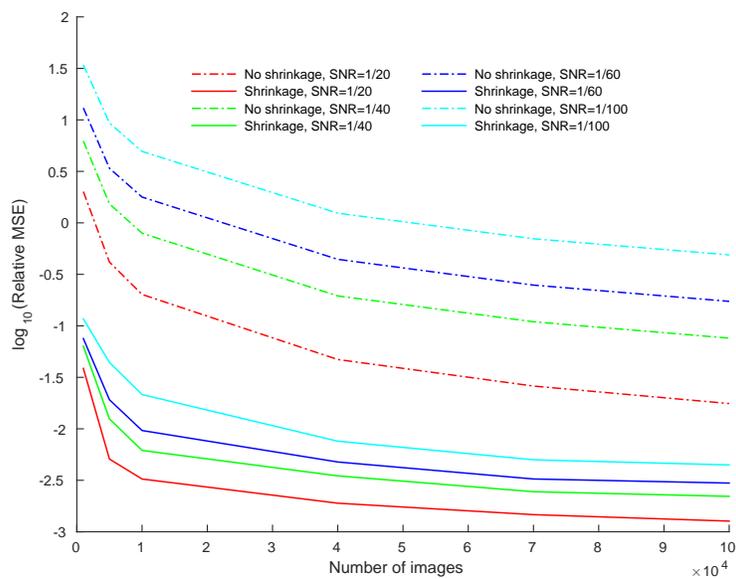}

\caption{\textbf{Relative MSE of the estimated covariance versus the number of images}: 
The 
relative MSE of the estimated covariance $\hat\Sigma$, with and without using eigenvalue shrinkage, as a function of number 
of images, for synthetic 
data 
generated using EMDB-6454.}
\label{fig:shrinkage}
\end{figure}

The importance of the eigenvalue shrinkage procedure is elucidated in Figure \ref{fig:shrinkage}.\ Here,
we compare the error in the estimated covariance with and without eigenvalue shrinkage, for 
varying number of images used in the experiment. The relative MSE of the estimated covariance $\hat{\Sigma}$ is defined as

\begin{equation}
 MSE_{rel} = \frac{||\Sigma-\hat{\Sigma}||_F^2}{||\Sigma||_F^2}
\end{equation}

\clearpage

\subsection{Simulated Noisy Dataset with Colored Noise}
\label{sec:colnoise}
\begin{figure}[]
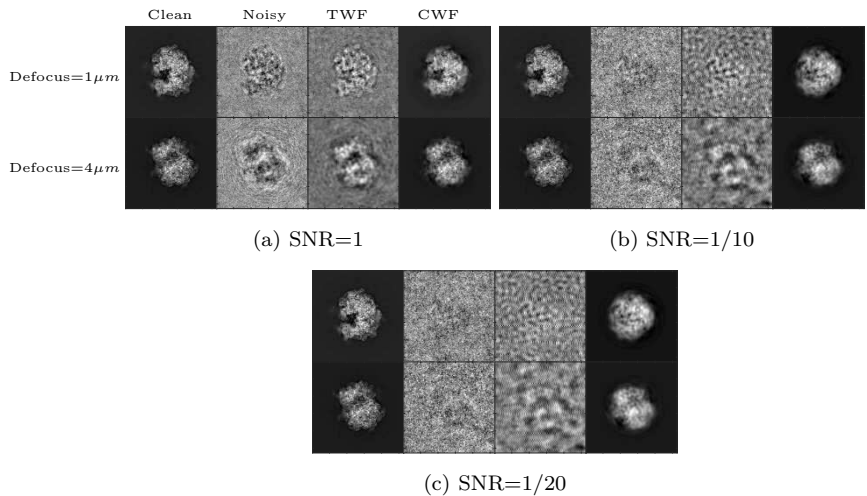

\centering
\subfloat[SNR=$1$]{\begin{overpic}[width=0.4\linewidth]{compare_ims_col_snr1by1_6454.eps}
\put(6,52){\tiny Clean}
\put(32,52){\tiny Noisy}
\put(55,52){\tiny TWF}
\put(80,52){\tiny CWF}
\put(-32,35){\tiny Defocus=$1\mu m$}
\put(-32,10){\tiny Defocus=$4\mu m$}
\end{overpic}%
}
\vspace{-1mm}
\subfloat[SNR=$1/10$]{\begin{overpic}[width=0.4\linewidth]{compare_ims_col_snr1by10_6454.eps}%
\end{overpic}%
\label{}}
\vspace{-1mm}
\subfloat[SNR=$1/20$]{\begin{overpic}[width=0.4\linewidth]{compare_ims_col_snr1by20_6454.eps}%
\end{overpic}%
\label{}}

\caption{\textbf{Synthetic colored noise}: Denoising results of 
CWF for the 
synthetic dataset with additive colored Gaussian noise, prepared from EMDB-6454, 
the P. falciparum 80S ribosome bound 
to E-tRNA, as detailed in the caption of Figure \ref{fig:ims_6454}.}
\label{fig:ims_6454_colored}
\end{figure}

The noise that corrupts images in cryo-EM is not perfectly white, but often 
colored. To simulate this, we perform experiments with synthetic data generated
from EMDB-6454 as described in \ref{sec:whitenoise}, this
time adding colored Gaussian noise with the noise
response $f(\text{k})=\frac{1}{\sqrt{(1+\text{k}^2)}}$ ($\text{k}$ is the radial 
frequency) to each 
clean,
CTF-affected projection image. Figure \ref{fig:ims_6454_colored}
shows the denoised images for this case.
%

\clearpage
\subsection{Experimental Dataset - TRPV1}
\label{sec:trpv1}
\begin{figure}[h]
\centering
{\begin{overpic}[width=0.8\textwidth]{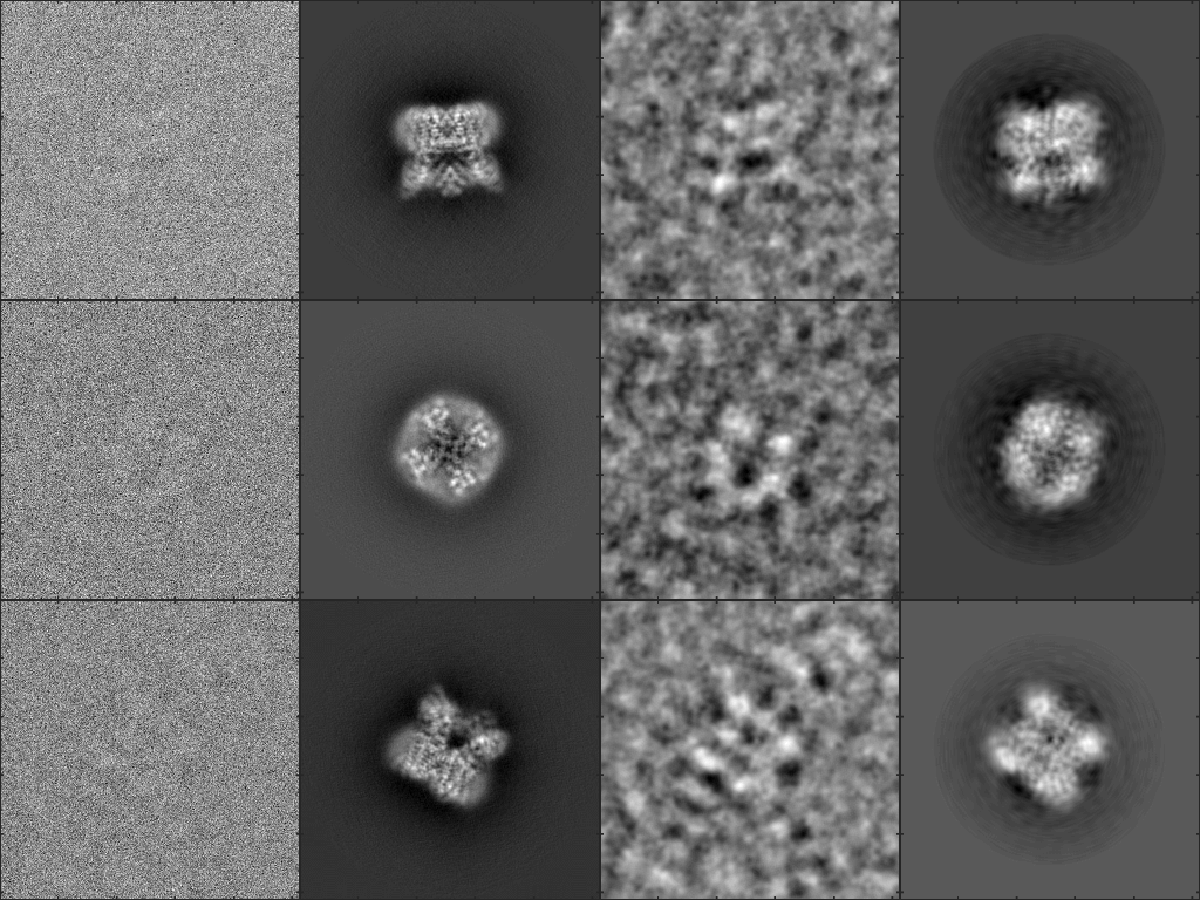}%
\put(12,77){\tiny Raw}
\put(27,77){\tiny Closest projection}
\put(59,77){\tiny TWF}
\put(84,77){\tiny CWF}
\end{overpic}
\label{}}
\caption{\textbf{Denoising an experimental dataset of TRPV1 \cite{trpv1_nature}}: 
Here we show, for three images in the dataset, the raw image, the closest true projection
image generated from the 3D reconstruction of the molecule (EMDB 5778),
the denoised image obtained using 
TWF, and the denoised image 
obtained using CWF. In this 
experiment, $35645$ images of size 256$\times$256
belonging to $935$ defocus groups were used. The amplitude contrast is $10\%$, 
the spherical aberration is $2$mm, and the voltage
is $300$kV. }
\label{fig:trpv1}
\end{figure}

We apply CWF to an experimental dataset of the TRPV1 ion channel, taken using a 
K2 direct 
electron detector. It is available on 
the public database Electron Microscope
Pilot Image Archive (EMPIAR) as EMPIAR-10005, and the 3D reconstruction is 
available on EMDB as EMDB-5778, courtesy of Liao et al. 
\cite{trpv1_nature}. The 
dataset consists of 35645 motion corrected, picked particle images of size $256 \times 256$ pixels
with a pixel size of $1.2156\AA$. Using 20 cores, calculating the Fourier Bessel
coefficients took 312 seconds while covariance estimation and Wiener filtering together
took 574 seconds. The result is shown in Figure \ref{fig:trpv1}. CWF retains 384
eigenvalues of $\Sigma$.
\vspace{3 mm}
 
\subsection{Experimental Dataset - 80S ribosome}
\begin{figure}[h]
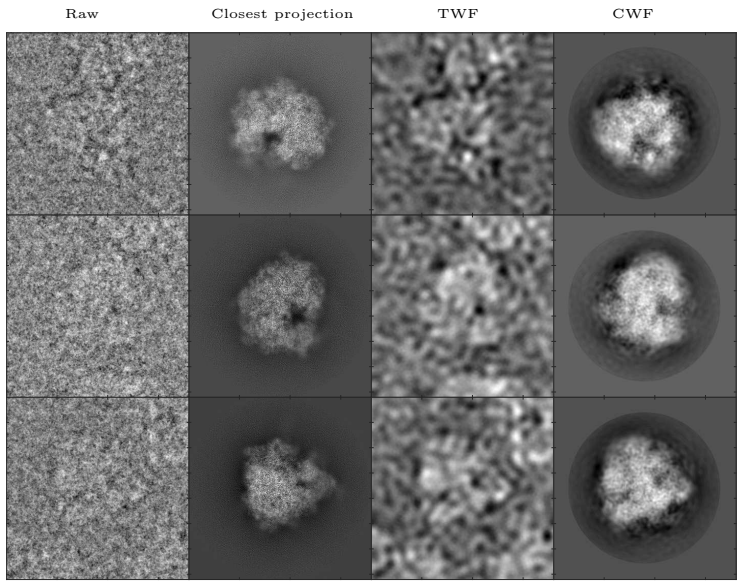

\centering
{\begin{overpic}[width=0.8\textwidth]{jsb_fig_80s.eps}%
\put(8,77){\tiny Raw}
\put(28,77){\tiny Closest projection}
\put(59,77){\tiny TWF}
\put(83,77){\tiny CWF}
\end{overpic}
\label{}}
\caption{\textbf{Denoising an experimental dataset of the 80S ribosome \cite{80s_bai}}: Here 
we 
show, for three images in the dataset, the raw image, the closest true projection
image generated from the 3D reconstruction of the molecule (EMDB 2660),
the denoised image obtained using 
TWF, and the denoised image 
obtained using CWF. In this 
experiment, the first $30000$ images out of the
105247 images in the dataset were used for covariance estimation.
The images are of size 360$\times$360 and
belong to $290$ defocus groups. The amplitude contrast is $10\%$, 
the spherical aberration is $2$mm, and the voltage
is $300$kV. }
\label{fig:real80s}
\end{figure}

We apply CWF to an experimental dataset of the Plasmodium falciparum 80S ribosome bound to the anti-protozoan drug emetine, taken 
using a FEI FALCON II 4k $\times$ 4k direct 
electron detector. The raw micrographs and picked particles are available on 
the public database EMPIAR as EMPIAR-10028, and the 3D reconstruction is 
available on EMDB as EMDB-2660, courtesy of Wong et al. 
\cite{80s_bai}. The 
dataset we used was provided by Dr. Sjors Scheres, and consists of 105247 motion 
corrected, picked particle images of size $360 \times 360$
with a pixel size of $1.34\AA$. Using 20 cores, calculating the Fourier Bessel
coefficients took 731 seconds while covariance estimation and Wiener filtering together
took 385 seconds. The result is shown in Figure \ref{fig:real80s}. CWF retains 962
eigenvalues of $\Sigma$.

\vspace{3 mm}

\subsection{Experimental Dataset - IP\textsubscript{3}R1}
\begin{figure}[h]
\centering
{\begin{overpic}[width=0.8\textwidth]{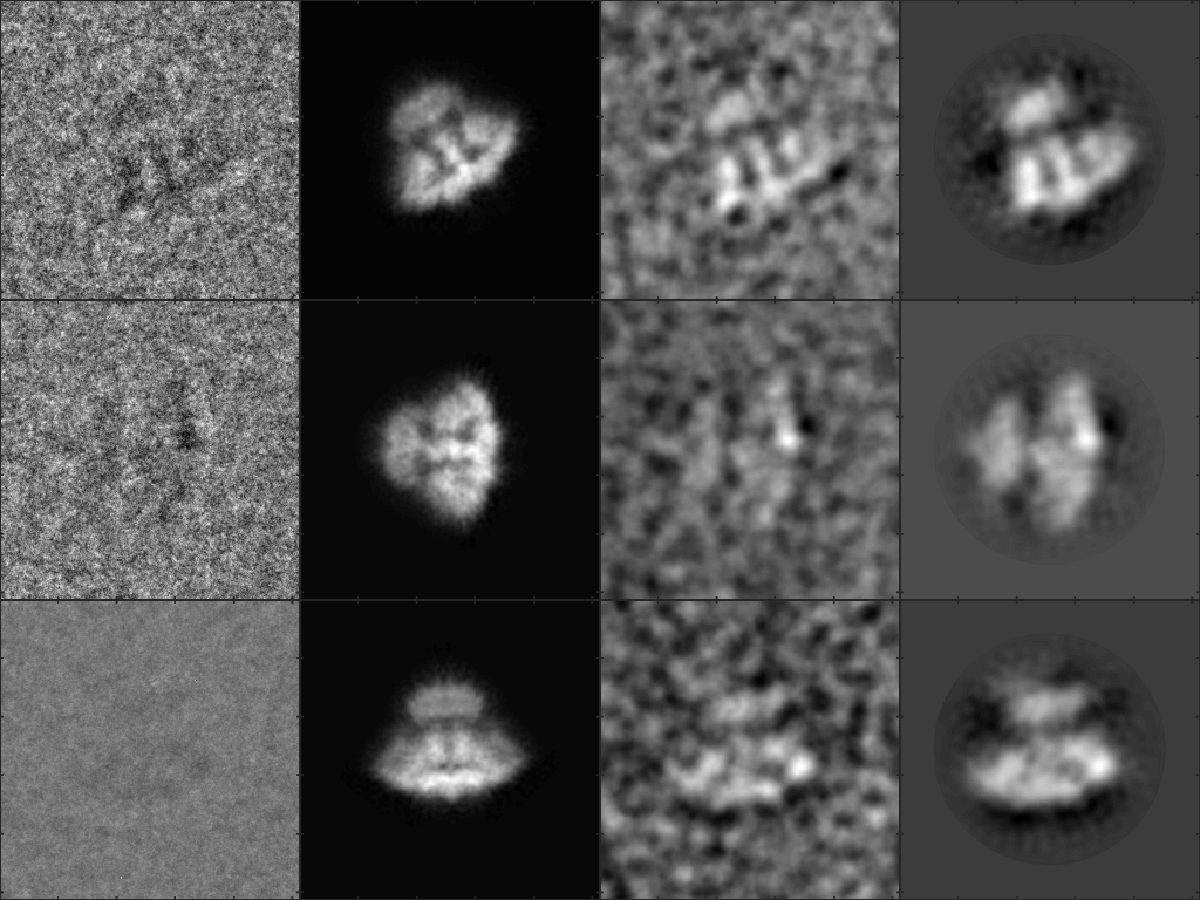}%
\put(10,77){\tiny Raw}
\put(27,77){\tiny Closest projection}
\put(57,77){\tiny TWF}
\put(83,77){\tiny CWF}
\end{overpic}
\label{}}
\caption{\textbf{Denoising an experimental dataset of IP\textsubscript{3}R1 \cite{ip3_paper}}:  Here we show, for 
three images in the dataset, the raw image, the closest true projection
image generated from the 3D reconstruction of the molecule (EMDB 5278), the denoised image obtained using 
TWF, and the 
denoised image obtained using 
CWF. In this 
experiment, $37382$ images of size $256\times256$
belonging to $851$ defocus groups were used. The amplitude contrast is $15\%$, 
the spherical aberration is $2$mm, and the voltage
is $200$kV. }
\label{fig:ip3}
\end{figure}

We apply CWF to an experimental dataset of the Inositol 1, 4, 5-triphosphate 
receptor 1 (IP\textsubscript{3}R1)
provided by Dr. Irina Serysheva, obtained using the older Gatan 4k $\times$ 4k 
CCD camera \cite{ip3_paper}. The 3D reconstruction obtained
from this dataset is available on EMDB as EMDB-5278. The dataset consists of 
37382 images of size 256$\times$256 pixels
with a pixel size of $1.81\AA$. Using 20 cores, calculating the Fourier Bessel
coefficients took 429 seconds while covariance estimation and Wiener filtering together
took 589 seconds. The result is shown in Figure \ref{fig:ip3}. CWF retains 290
eigenvalues of $\Sigma$.
\vspace{3 mm}

\subsection{Experimental Dataset - 70S ribosome}
\begin{figure}[h]
\centering
{\begin{overpic}[width=0.8\textwidth]{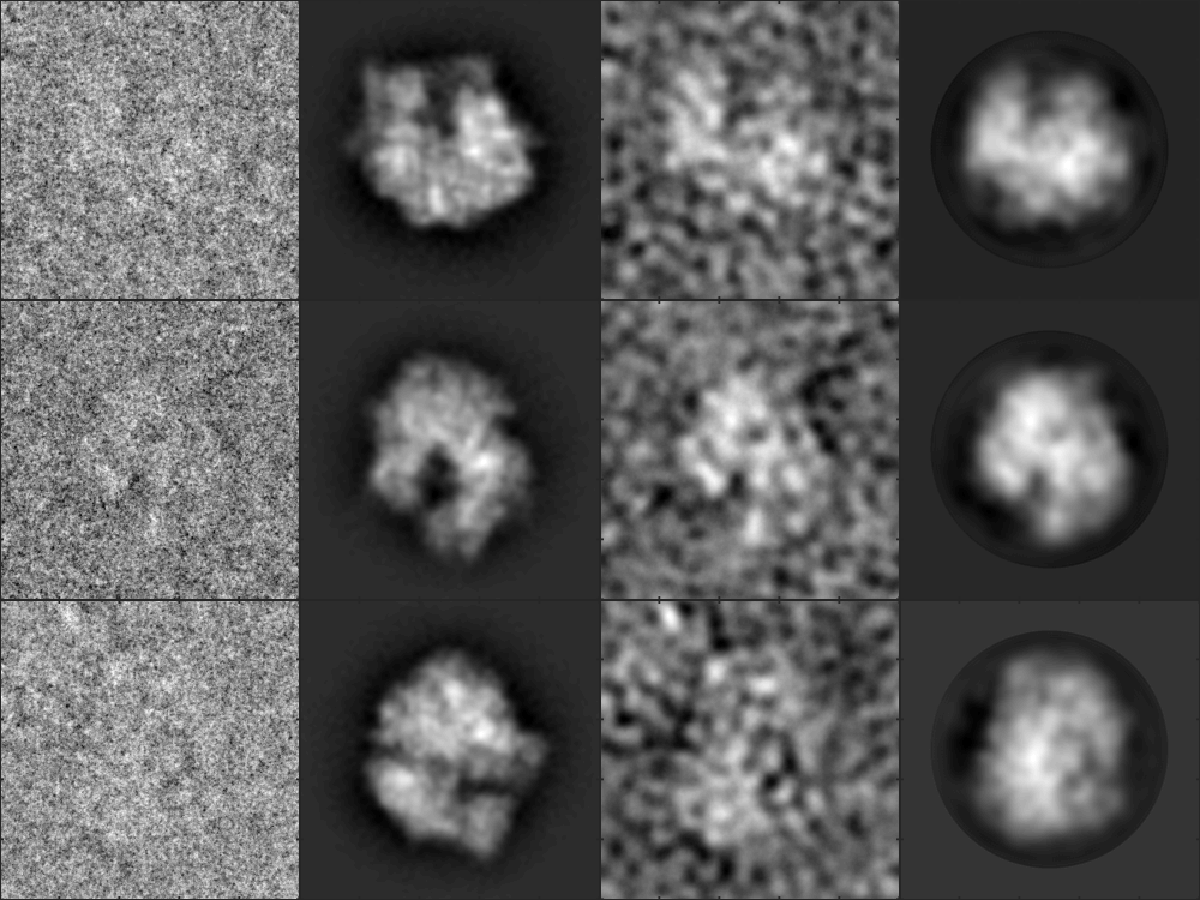}%
\put(10,77){\tiny Raw}
\put(28,77){\tiny Closest projection}
\put(58,77){\tiny TWF}
\put(82,77){\tiny CWF}
\end{overpic}
\label{}}
\caption{\textbf{Denoising an experimental dataset of 70S \cite{70s_frank}}: Here we show, for 
three images in the dataset, the raw image, the closest true projection
image generated from the 3D reconstruction of the molecule (EMDB 5360),
the denoised image obtained using 
TWF, and the denoised image obtained using 
CWF. In this 
experiment, the first $99979$ images out
of the $216517$ images in the dataset were used for covariance estimation. The images are of size 250$\times$250
and belong to $38$ defocus groups. The amplitude contrast is $10\%$, 
the spherical aberration is $2.26$mm, and the voltage
is $300$kV. }
\label{fig:real70s}
\end{figure}

We apply CWF to an experimental dataset of the 70S ribosome
provided by Dr. Joachim Frank's group \cite{70s_frank}. This heterogeneous 
dataset consists of $216517$ images
of size $250\times250$ pixels
with a pixel size of $1.5\AA$, obtained
using the older TVIPS TEMCAM-F415 (4k x 4k) CCD detector. The 3D reconstruction obtained
from this dataset is available on EMDB as EMDB-5360. Using 20 cores, calculating the Fourier Bessel
coefficients took 1174 seconds while covariance estimation and Wiener filtering together
took 113 seconds. The result is shown in 
Figure \ref{fig:real70s}. CWF retains 219
eigenvalues of $\Sigma$.
\vspace{3 mm}

\subsection{Outlier Detection}

In the cryo-EM pipeline, a significant amount of time is spent on discarding outliers by visual inspection
after the particle picking step. CWF provides an automatic way to classify picked particles
into ``good'' particles and outliers. The classifier uses the contrast of a denoised image to
determine if it is an outlier.

The specimen particles can be at various depths in the ice layer at the time of imaging, so the acquired
projection images can have different contrasts. The contrast can be modeled
as an additional scalar parameter $\alpha$ for each acquired noisy projection image as in eqn.\ \ref{eqn:contrast},
typically as a uniformly distributed random variable spread about its mean at 1.

\begin{equation}
 Y_i = \alpha_i A_i X_i + \xi_i, \quad i=1,2,\ldots,n
\label{eqn:contrast}
\end{equation}

\begin{figure}[]
\centering
\subfloat[]{\begin{overpic}[width=0.48\linewidth]{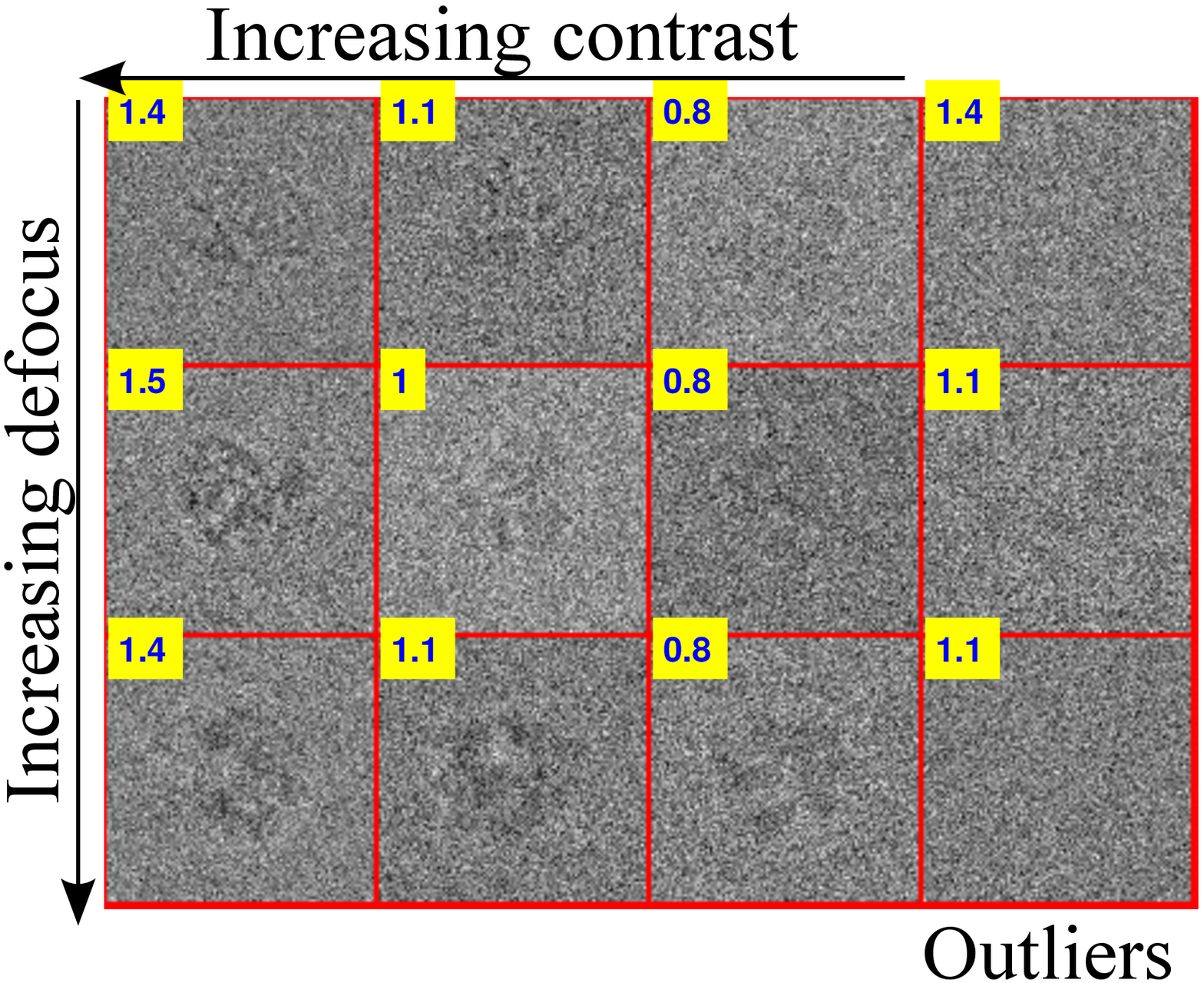}%
\end{overpic}%
\label{fig:raw_outlier}}
\quad
\subfloat[]{\begin{overpic}[width=0.48\linewidth]{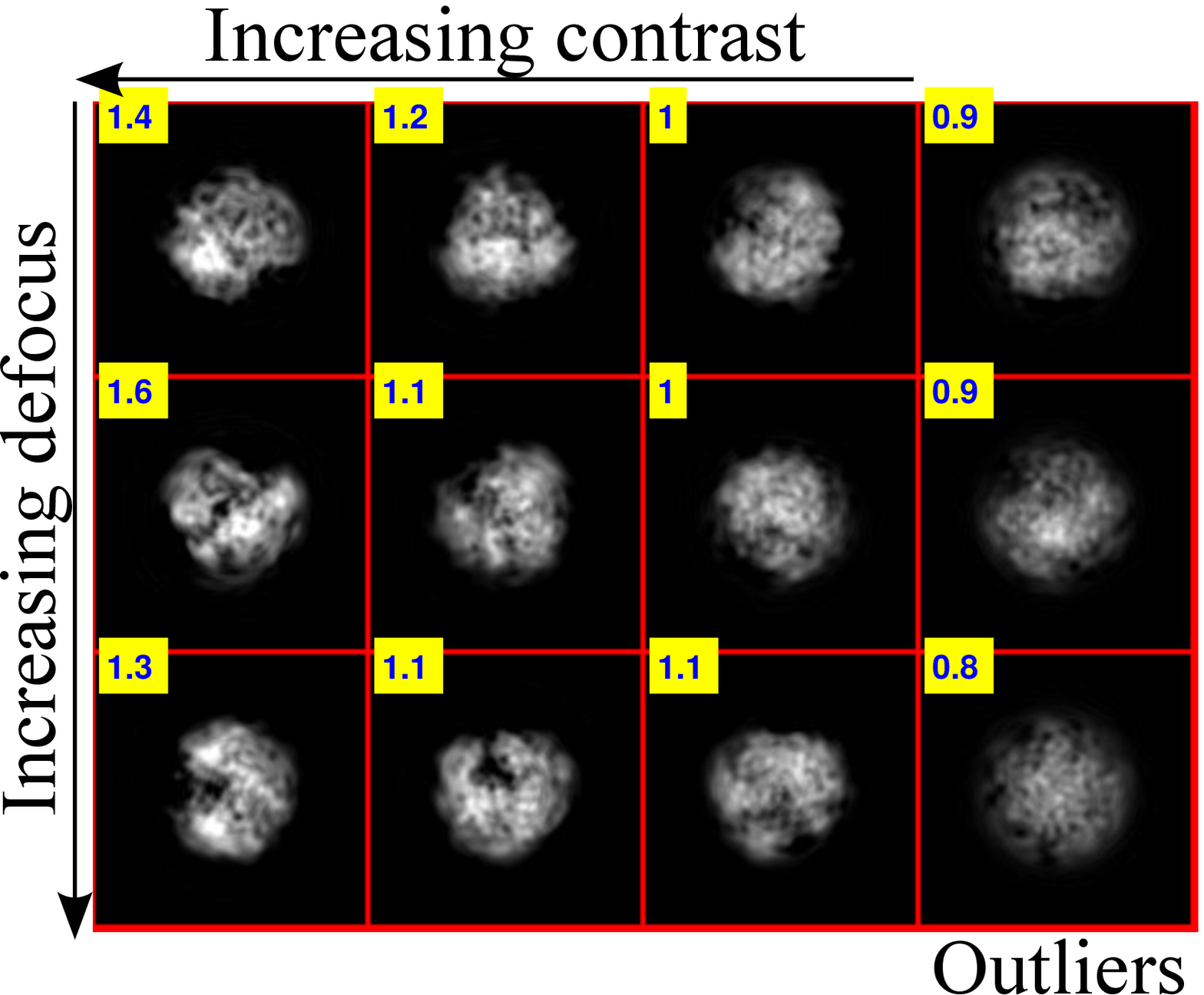}
\end{overpic}
\label{fig:den_outlier}} \\
\subfloat[]{\begin{overpic}[width=0.125\linewidth]{mean_image.eps}%
\end{overpic}%
\label{fig:mean_image}}
\quad
\subfloat[]{\begin{overpic}[width=0.7\linewidth]{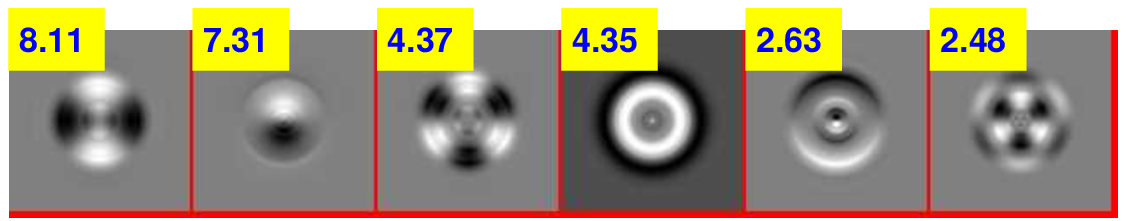}
\end{overpic}
\label{fig:eigenims}}
\caption{(a) \textbf{Raw images}: 
A sample of synthetic 
data 
generated using EMDB-6454 with additive colored Gaussian noise at SNR=1/20.\ $10\%$ of
the projection images are replaced by pure noise. The contrast parameter $\alpha$ ranges from 0.75 to 1.5. The outliers are shown in the last column. Inset 
in a yellow box is the contrast of each image.
(b) \textbf{Denoised images}: 
The denoised images using CWF. Notice the low contrast outliers in the last column. (c) \textbf{Estimated Mean Image}
(d) \textbf{Top 6 eigenimages}: 
Inset in a yellow box is the corresponding eigenvalue.}
\end{figure}

We absorb the contrast $\alpha$ into $\textbf{X}$ and estimate $\alpha_i X_i$ in this case, using the same procedure as before.
We perform an experiment with synthetic data generated using EMDB-6454 with additive colored Gaussian noise at SNR=1/20, and $\alpha \in [0.75,1.5]$.\ $10\%$ of the
projection images are replaced by ``outliers'', that is, pure noise images containing no signal. 
Fig. \ref{fig:mean_image} shows the estimated mean image $\mu$, and Fig. \ref{fig:eigenims} shows the top 6 principal components
of the estimated covariance $\hat\Sigma$, also known as eigenimages. Fig. \ref{fig:raw_outlier} and Fig. \ref{fig:den_outlier} show 
a sample of raw and denoised images respectively. High contrast images enjoy a higher 
SNR and are thus of interest for subsequent steps of the pipeline. On the other hand, outlier images, which typically have low contrast
after denoising, can be automatically detected by a linear classifier
after CWF and discarded from the dataset. In the experiment shown in Fig. \ref{fig:raw_outlier} and \ref{fig:den_outlier}, a classifier
with a threshold of $0.95$ for the contrast discards $95\%$ of the outliers, while $3\%$ of the inliers are also discarded
in the process.

One can also use a different classifier based on features like the relative energy of the image before and after denoising, etc.
However, outliers that look like particles, for example, images belonging to a different class of a heterogeneous dataset which act as ``contaminants'',
are difficult to detect using this method.

\section{Conclusion}
In this paper we presented a new approach for image restoration of cryo-EM images, CWF, whose
main algorithmic components are covariance estimation and deconvolution using Wiener filtering.
CWF performs both CTF correction, by correcting the Fourier phases and amplitudes of the images, 
as well as denoising, by eliminating the noise thereby improving the SNR of the resulting images.
In particular, since CWF applies Wiener filtering in the data-dependent basis of principal components (``eigenimages"),
while TWF applies Wiener filtering in the data-independent Fourier basis, we see
in numerical experiments that CWF performs better than TWF, and considerably better at high noise levels.
We demonstrated the ability of CWF to restore images for several 
experimental datasets, acquired with both CCD detectors and the state-of-the-art direct electron detectors.

Due to the high noise level typical in cryo-EM images, 2D classification is performed before estimating a 3D ab-initio model. Class averages 
enjoy a higher SNR and are used to estimate viewing angles and obtain an initial model.
For future work, it remains to be seen whether the resulting denoised images from CWF can be directly used to estimate viewing angles,
without performing classification and averaging.
Another possible future direction is integration of CWF into existing 2D class averaging procedures in order to improve their performance.

\section{Acknowledgements}
We are thankful to Maofu Liao, Sjors Scheres, Irina Serysheva and Joachim Frank for generously
providing us with the experimental datasets. We thank Xiaochen Bai for answering our questions about the 80S dataset.
We are grateful to Yoel Shkolniskly and Zhizhen Zhao for help with the code. We also thank Zhizhen Zhao for reviewing earlier 
versions of this manuscript, and for numerically expressing isotropic linear operators such as the CTF in the Fourier-Bessel basis.
We thank Fred Sigworth and Joakim And\'en for many
helpful discussions about this work. We also thank Joakim And\'en for pointing out the symmetrization required for the linear system to be solved by conjugate gradient.   
We are grateful to both reviewers and the editor
for their helpful comments. The authors were partially supported by Award Number R01GM090200 from the NIGMS,
FA9550-12-1-0317 and FA9550-13-1-0076 from AFOSR, LTR DTD 06-05-2012 from the Simons
Foundation, and the Moore Foundation Data-Driven Discovery Investigator Award.

\clearpage

\bibliography{mybibfile}

 \end{document}